\documentclass{article}

\usepackage[final]{corl_2026}

\usepackage{algorithm}
\usepackage{algpseudocode}
\usepackage{graphicx}
\usepackage{booktabs}
\usepackage{amsmath}
\usepackage{amssymb}
\usepackage{pifont}
\usepackage{wrapfig}
\usepackage{makecell}
\usepackage{caption}
\usepackage{xspace}
\usepackage{multirow}

\usepackage{colortbl}

\definecolor{MetaGreen}{rgb}{0.92,0.97,0.94}
\newcommand{\oursrow}{\rowcolor{MetaGreen}}

\newcommand{\eg}{\textit{e.g.},\xspace}

\title{From Video Frames to Robot Meta-Frames:
A Unified Latent Interface for World Action Models}

\author{
  Guoqiang Liang$^{1,2}$ \quad
  Yiqi Lin$^{1,2}$ \quad
  Yanzhe Chen$^{1}$ \quad
  Kevin Yuchen Ma$^{1}$ \quad
  Mike Zheng Shou$^{1}$\thanks{Corresponding author.} \\
  $^{1}$National University of Singapore \\
  $^{2}$National University of Singapore Guangzhou Research Translation and Innovation Institute, \\
  Guangzhou, Guangdong 510700, China
}

\begin{document}
\maketitle

\begin{abstract}
Pretrained video diffusion models (VDMs) offer a powerful foundation for World Action Models (WAMs), which jointly predict robot actions and future observations.
Building such models requires injecting low-dimensional robot signals (\eg proprioception, actions, and values) into a latent space shaped by natural videos, and a key design choice is the interface through which these signals enter: existing schemes treat it as a formatting detail, spending a separate latent frame on each signal or bolting on dedicated encoders and decoders.
We argue that signal injection is instead a distribution-alignment problem:
how well a representation matches the prior of the pretrained VDMs during denoising.
Using the relative reconstruction error of a frozen VDM denoiser as a probe, we find that conventional spatial encodings sit far off this prior, whereas frequency-domain packing yields latents that the denoiser corrects far less.
Motivated by this diagnostic result, we propose MetaWAM, which folds all robot signals into a single unified meta-frame: each signal is assigned to a disjoint frequency-domain channel band and lifted to the spatial latent by an inverse DCT, yielding a decodable interface aligned with the pretrained VDM prior without any added module or architectural change.
A fixed band layout together with slot-aware routing lets the same meta-frame support policy, world-model, and value prediction over a shared set of robot variables.
On RoboCasa, MetaWAM reaches 68.9\% success while collapsing the per-signal frames into one and cutting inference latency by 39\%, and attains 85.0\% success on three real-world Franka tasks.
Code and models are available at \url{https://github.com/showlab/MetaWAM}.
\end{abstract}

\section{Introduction}
Vision-Language-Action models have become a powerful paradigm for generalist robot
control, yet they learn reactive observation-to-action mappings without modeling how the world
evolves under their actions~\citep{wang2026world}.
\emph{World Action Models} (WAMs) close this gap by adapting pretrained video diffusion models
(VDMs), which carry strong spatiotemporal priors over scene dynamics, to predict actions and
future observations jointly.
Existing WAMs differ mainly in how they couple world prediction and control. \emph{Cascaded}
WAMs use the video model as a learned simulator and recover actions through a separate
inverse-dynamics model or policy~\citep{unipi2023, genie2024, liu2026world}, while \emph{Joint} WAMs let one
generative backbone emit actions, future observations, and values within a single denoising
process~\citep{cosmospolicy2026,dreamzero2026}. Across these designs, a key interface is often
treated as a formatting detail: how low-dimensional robot variables, such as proprioception,
actions, task type, and values, should enter a latent space shaped by natural-video statistics.
This raises the question: \emph{how should structured robot signals be represented so that a
pretrained VDM treats them as in-distribution during denoising?}

\begin{figure*}[!t]
\centering
    \includegraphics[width=1\linewidth]{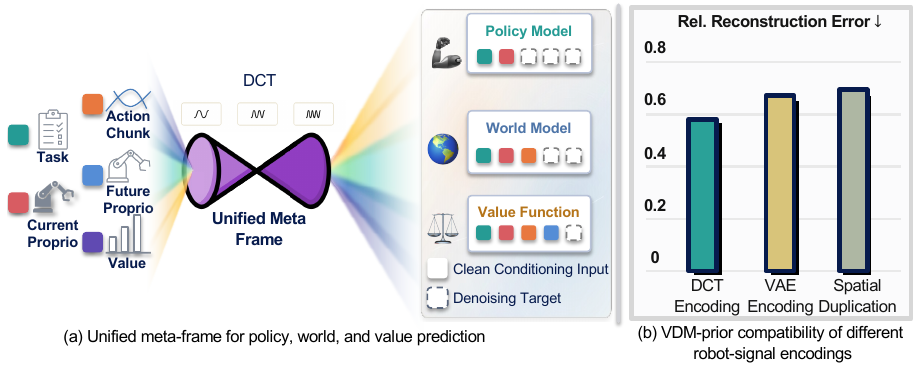}
    \caption{\textbf{MetaWAM packs robot signals into one distribution-aware meta-frame.}
\textbf{(a)} Disjoint DCT bands route signals as inputs or denoising targets for policy,
world-model, and value prediction.
\textbf{(b)} Frozen-denoiser relative reconstruction error at $\sigma=1.0$ ($\downarrow$):
Spatial Duplication (\eg Cosmos Policy~\citep{cosmospolicy2026}) and VAE Encoding
(\eg AgiBot-World~\citep{bu2025agibot}) require larger corrections, while our
DCT Encoding is corrected least.}
    \label{fig:teaser}
\centering
\end{figure*}

A straightforward solution injects each robot signal as its own video-like latent frame,
either by spatially tiling the raw signal across the frame~\citep{cosmospolicy2026} or by
rendering it into an image and re-encoding it through the pretrained VAE~\citep{bu2025agibot};
both preserve backbone compatibility but inflate the sequence length and the quadratic attention
cost it incurs. Other designs attach dedicated action encoders or decoders~\citep{dreamzero2026},
improving action modeling at the cost of modifying the pretrained backbone. We view signal
injection as a \emph{distribution-alignment} problem: a useful representation must remain
compatible with the VDM prior seen by the pretrained denoiser. As previewed in
Fig.~\ref{fig:teaser}(b), frequency-domain encoding produces latents that the frozen denoiser
corrects far less than spatial encodings.

To examine this design space, we run diagnostic experiments that trace candidate injections through
VDM denoising, covering both single-signal injection and multi-signal packing, and measure how
strongly a frozen denoiser corrects the injected latent across noise levels. These experiments show
that the arrangement of robot signals in latent space measurably affects the denoiser:
frequency-domain packing yields latents it corrects far less than spatial tiling or discontinuous
canvas layouts. Compactness alone is therefore insufficient; an effective interface must be
\emph{distribution-aware}.

Motivated by these findings, we propose MetaWAM, which folds all heterogeneous robot signals into
a single unified \emph{meta-frame} (Fig.~\ref{fig:teaser}(a)). Rather than allocating a latent frame
to each signal, MetaWAM assigns every signal, including current and future proprioception, action
chunks, task type, and value, to a disjoint frequency-domain channel band within one frame, and
lifts the packed coefficients to the spatial latent via an analytical inverse DCT. This produces
smooth latent patterns well aligned with the pretrained denoiser, while a single meta-frame becomes
a multiplexed carrier for many low-dimensional signals. The design requires no additional action
encoder, learned decoder, or modification to the denoising architecture, and recovers each signal
analytically at readout.

We instantiate MetaWAM on a Cosmos-Predict2 backbone and evaluate it on RoboCasa, LIBERO,
LIBERO-Plus, and three real-world Franka tabletop tasks. MetaWAM exceeds the RoboCasa frame-injection baseline while collapsing
multiple robot-signal frames into one. Our contributions are:
(1) we recast robot-signal injection as a distribution-aware latent-interface problem for Joint WAMs;
(2) we provide diagnostic evidence that frequency-domain packing aligns better with a pretrained
VDM prior than spatial signal encodings; and
(3) we introduce a unified frequency-domain meta-frame that carries all robot signals in a single
decoder-free latent frame without architectural changes.

\section{Related Work}
\textbf{World Action Models.}
World Action Models (WAMs) recast robot control as generative modeling over future states and
actions~\citep{wang2026world}. Cascaded WAMs generate or represent future observations first and
then recover actions through a separate policy or inverse-dynamics model~\citep{unipi2023,
genie2024,jang2025dreamgen}. Joint WAMs instead couple action and future-state prediction inside
one backbone. Cosmos Policy~\citep{cosmospolicy2026} injects actions, proprioception, future
states, and values as additional latent frames, preserving the pretrained video architecture;
DreamZero~\citep{dreamzero2026} uses dedicated action encoders and decoders around a video
backbone; and UWM/UVA~\citep{uwm2025,uva2025} decouple video and action diffusion schedules to
enable mixed robot and video training. Recent efficiency-oriented WAMs reduce inference cost by
changing what is denoised or attended to, such as removing future-frame generation at test
time~\citep{fastwam2026} or emphasizing action-centered decoding~\citep{ye2026gigaworld}.
MetaWAM is orthogonal to these directions: it keeps the pretrained denoising architecture intact
but changes how robot signals are represented inside the latent sequence. Rather than treating
signal injection as a formatting choice, we study its compatibility with the pretrained VDM prior
and introduce a distribution-aware interface.

\textbf{Latent signal representations.}
A broad line of work studies how robot or agent signals should be encoded into compact latent
forms. Learned schemes include VQ-VAE latent action codes derived from visual
dynamics~\citep{lapa2024,moto2024}, world-model-grounded latent actions~\citep{lawm2025}, highly
compressed discrete action chunks~\citep{vqace2024}, and modality-specific latent positions for
proprioception, actions, and future tokens~\citep{flare2025}. Transform-based schemes exploit
frequency structure: FAST~\citep{fast2025} applies 1D DCT to robot action sequences, while
wavelet tokenizers~\citep{wavelettokenizer2024}, FA-VAE~\citep{favae2025}, and
DocPedia~\citep{docpedia2024} use frequency-domain representations for visual or document
modeling. MetaWAM differs in target and mechanism: it uses channel-wise 2D DCT bands to pack
heterogeneous robot signals directly into the latent space of a pretrained VDM, lifts them back to
the spatial domain by analytical iDCT, and recovers them without a learned decoder.

\section{Motivating Analysis}
\label{sec:motivating_analysis}
\begin{figure*}[!t]
\centering
    \includegraphics[width=1\linewidth]{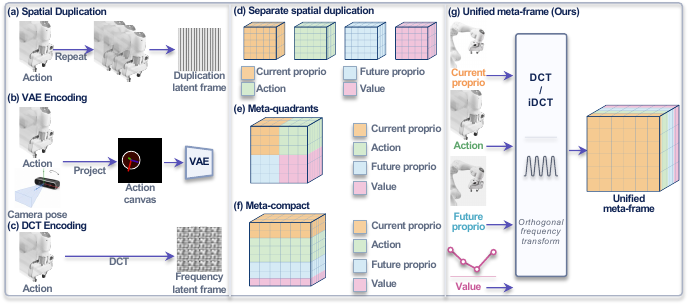}
    \caption{\textbf{Robot-signal injection schemes used in the diagnostic study.}
\textbf{(a--c)} Single-signal encodings compare Spatial Duplication, VAE Encoding, and DCT Encoding.
\textbf{(d--g)} Multi-signal packings compare Separate Spatial Duplication, Meta-Quadrants,
Meta-Compact, and our Unified Meta-Frame, which assigns signals to disjoint channel bands and
lifts them to one spatial latent by orthonormal DCT/iDCT.}
    \label{fig:motivation}
\centering
\vspace{-2mm}
\end{figure*}

\emph{What kind of robot-signal injection does a pretrained VDM actually accept?} We probe
candidate injections by a single criterion, VDM-prior compatibility, defined by the
relative reconstruction error of a frozen Cosmos-Predict2 denoiser~\citep{cosmoswfm2025} on the injected latent:
the denoising residual is normalized by the magnitude of the clean latent. Reference latents are
drawn from real RoboCasa observations, and we evaluate this error across noise levels $\sigma$.
An in-distribution latent needs only small relative corrections; a large residual marks it as
off-prior. Fig.~\ref{fig:motivation} and Tab.~\ref{tab:vdm_prior_compat} summarize the
schemes and results.

\textbf{Single-signal injection.}
We isolate a single signal and compare the three schemes in Fig.~\ref{fig:motivation}(a--c):
\emph{Spatial Duplication} (the Cosmos Policy~\citep{cosmospolicy2026} strategy), \emph{VAE Encoding} (canvas overlay
through the VAE~\citep{bu2025agibot}), and \emph{DCT Encoding} (raw DCT coefficients written into a latent frame). Tab.~\ref{tab:vdm_prior_compat}(a) reports the denoiser's
relative reconstruction error at every $\sigma$. Spatial Duplication, the simplest backbone-compatible
choice, incurs the largest corrections: constant row-wise tiling is precisely the pattern a
video-trained denoiser pushes away from. VAE Encoding fares no better; although designed to pass cleanly through the VAE, it is still treated as off-prior by the denoiser. DCT Encoding instead
spreads the signal into smooth cosine bases and sits closest to the prior, attaining the lowest
relative reconstruction error at every $\sigma\!\le\!5.0$. What an effective injection must target is
compatibility with the denoiser, not merely the encoder a latent passes through.

\textbf{Multi-signal packing.}
We use the corresponding action chunks, proprioceptive states, and value labels from the
same RoboCasa samples to compare four layouts (Fig.~\ref{fig:motivation}(d--g)).
\emph{Separate Spatial Duplication} spatially duplicates each signal into its own latent
frame, using three frames in total; its reported error is averaged over these frames.
\emph{Meta-Quadrants} puts all three signals into one frame, assigning each an equally sized
spatial region and replicating its values along the channel dimension within that region.
\emph{Meta-Compact} also uses one frame and channel-wise replication, but allocates spatial
regions according to signal dimensionality: action chunks occupy a larger region, while
proprioception and value occupy smaller ones. These two layouts therefore test whether
spatial consolidation alone, with either equal or dimension-dependent allocation, is sufficient.
Our \emph{Unified Meta-Frame} instead assigns signals to disjoint channel bands, writes their
values into fixed DCT coefficient slots, and applies a per-channel 2D inverse DCT to form
one spatial latent. Signals are separated by channel allocation rather than spatial boundaries.

Table~\ref{tab:vdm_prior_compat}(b) shows that this choice matters beyond frame count.
At $\sigma=1.0$, Meta-Quadrants and Meta-Compact incur relative reconstruction errors of
0.6730 and 0.6234, respectively, compared with 0.5613 for our Unified Meta-Frame and
0.6178 for Separate Spatial Duplication. Our layout achieves the lowest error at every tested
noise level, combining a single-frame representation with greater frozen-denoiser compatibility.

\begin{table}[t]
\centering
\begin{minipage}[t]{0.49\linewidth}
\centering
\label{tab:exp_single}
\resizebox{\linewidth}{!}{%
\setlength{\tabcolsep}{4pt}
\renewcommand{\arraystretch}{1.1}
\begin{tabular}{lccccc}
\toprule
\textbf{Method} & 0.01 & 0.05 & 0.2 & 1.0 & 5.0 \\
\midrule
Spatial Duplication & 0.0141 & 0.0672 & 0.2335 & 0.6459 & 0.8953 \\
VAE Encoding        & 0.0147 & 0.0699 & 0.2388 & 0.6379 & 0.8926 \\
\oursrow DCT Encoding        & \textbf{0.0133} & \textbf{0.0633} & \textbf{0.2199} & \textbf{0.5922} & \textbf{0.8646} \\
\bottomrule
\end{tabular}%
}
\caption*{{(a) Single-signal injection.}}
\end{minipage}
\hfill
\begin{minipage}[t]{0.49\linewidth}
\centering
\label{tab:exp_multi}
\resizebox{\linewidth}{!}{%
\setlength{\tabcolsep}{4pt}
\renewcommand{\arraystretch}{1.1}
\begin{tabular}{lccccc}
\toprule
\textbf{Method} & 0.01 & 0.05 & 0.2 & 1.0 & 5.0 \\
\midrule
Sep.\ Spatial Dup.$^\dagger$ & 0.0142 & 0.0677 & 0.2336 & 0.6178 & 0.8677 \\
Meta-Quadrants               & 0.0147 & 0.0705 & 0.2471 & 0.6730 & 0.9291 \\
Meta-Compact                 & 0.0132 & 0.0622 & 0.2160 & 0.6234 & 0.9392 \\
\oursrow Unified Meta-Frame         & \textbf{0.0126} & \textbf{0.0598} & \textbf{0.2061} & \textbf{0.5613} & \textbf{0.8159} \\
\bottomrule
\end{tabular}%
}
\caption*{{(b) Multi-signal packing.}}
\end{minipage}
\caption{VDM-prior compatibility, measured as the relative reconstruction error of a frozen
Cosmos-Predict2~\citep{cosmoswfm2025} at noise level $\sigma$ ($\downarrow$ lower is better).
\textbf{Bold} marks the best method per $\sigma$.
$^\dagger$Separate Spatial Duplication is averaged over 3 frames.}
\label{tab:vdm_prior_compat}
\end{table}

\textbf{Takeaway.}
Two principles follow. First, injection must be \emph{distribution-aware}, since compactness alone
does not guarantee compatibility with the VDM's natural-video prior. Second, multi-signal WAMs
should not spend a full latent frame per low-dimensional signal.
MetaWAM further adopts a
\emph{decoder-free} design, recovering signals analytically so that the savings from fewer
frames are not offset by added modules. The method below realizes these choices as a unified
frequency-domain meta-frame with slot-aware task routing. \textit{Further analysis on the
Wan-based~\citep{wan2025} backbone is provided in the supplementary material.}

\section{Method}
\label{sec:method}
\begin{figure*}[!t]
\centering
    \includegraphics[width=1\linewidth]{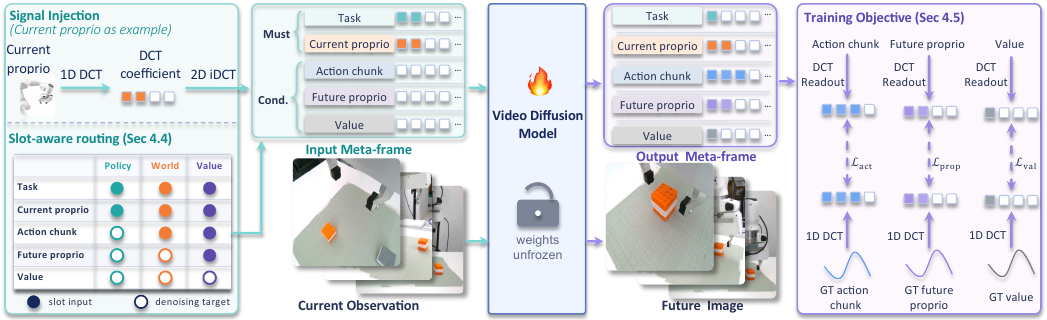}
    \caption{{\textbf{Overview of MetaWAM.} Normalized signals are packed into fixed coefficient
slots and lifted by 2D inverse DCT. Slot-aware masks select clean inputs and prediction targets
for policy, world-model, and value learning.}}
    \label{fig:framework}
\centering
\end{figure*}

We introduce MetaWAM, a World Action Model built around a \emph{unified robot-signal latent
interface} (Fig.~\ref{fig:framework}).
Rather than design a new backbone, we rethink how
structured robot variables enter an existing pretrained VDM, replacing frame-wise injection with
a pair of compact \emph{meta-frames}: an input meta-frame for robot-state conditioning and an
output meta-frame for action, future-state, and value prediction. Each meta-frame exposes its
variables not as raw spatial patterns but through structured frequency bands, turning one frame
into a multiplexed carrier for many low-dimensional signals while preserving the denoising architecture and removing redundant latent frames.

\subsection{Overview: A Unified Robot-Signal Interface}
Let a pretrained VDM operate on a latent video sequence
$\mathbf{z}\in\mathbb{R}^{C\times T\times H\times W}$, where each temporal slot is a spatial
latent frame. Frame-injection WAMs place each robot variable into its own additional frame,
forcing every scalar signal to occupy an image-sized frame and inflating $T$.
MetaWAM instead uses the following 8-frame latent sequence:
\begin{equation}
\mathbf{z}
=
(
z^{m}_{\mathrm{in}},
z^{w}_{t}, z^{p}_{t}, z^{s}_{t},
z^{m}_{\mathrm{out}},
z^{w}_{t+H}, z^{p}_{t+H}, z^{s}_{t+H}
),
\label{eq:latent_sequence}
\end{equation}
where $z^{m}_{\mathrm{in}}$ and $z^{m}_{\mathrm{out}}$ are the input and output meta-frames, and
$w,p,s$ index current and future observations from the wrist, primary, and secondary cameras.
The input meta-frame carries conditioning information, namely a one-hot task type and the current proprioception $\rho_t$, while the output meta-frame is the shared prediction interface
for action chunks, future proprioception, and value.

This consolidation shortens the sequence. A frame-injection baseline spends a separate frame on
each of input proprioception, action, future proprioception, and value; MetaWAM folds input
proprioception into the tokenizer's mandatory blank slot ($z^{m}_{\mathrm{in}}$) and merges the
three output signals into $z^{m}_{\mathrm{out}}$, yielding the 8-frame sequence of
Eq.~\eqref{eq:latent_sequence}: one conditioning meta-frame, six visual frames, one output
meta-frame. As self-attention scales quadratically in $T$, this cuts pairwise attention by
roughly $47\%$ with the architecture untouched.

\subsection{Distribution-Aware Spectral Lifting}
Broadcasting a robot vector over the spatial grid yields constant patterns; packing values into
spatial regions yields discontinuities, which are the failure modes diagnosed in Sec.~\ref{sec:motivating_analysis}. We therefore encode robot variables in the frequency domain
and lift them back to the spatial latent.

\textbf{Channel-band layout.}
For each meta-frame we build a sparse coefficient tensor $F\in\mathbb{R}^{C\times H\times W}$
whose \emph{disjoint channel bands} hold the distinct robot signals, one band per signal:
\begin{equation}
F = \big[\,F_{\text{task}}, F_{\rho_t}, F_{a}, F_{\rho_{t+H}}, F_{v}\,\big],
\label{eq:channel_bands}
\end{equation}
{where the bands form a predefined partition of the latent channels. Let $q$ collect the
one-hot task indicator and normalized robot signals. We construct $F=\operatorname{Pack}(q)$
by flattening or grouping each signal and directly writing its values into fixed coefficient
slots on the $k_H=0$ row of its channel band; all unused entries are zero. This packing
preserves the signal values without a temporal DCT or coefficient truncation.
The allocation is fixed across training and inference and introduces no learned parameters.}

\textbf{Spatial lifting.}
We obtain the spatial meta-frame $z^{m} = \mathcal{T}^{-1}_{\mathrm{2D}}(F)$ by applying an orthonormal 2D inverse DCT per channel.
{This distributes the packed coefficients over the $H\times W$ grid through the DCT basis. A spatial 2D DCT on the
denoised output meta-frame followed by fixed slot readout gives
\begin{equation}
\hat{F} = \mathcal{T}_{\mathrm{2D}}(\hat{z}^{m}),
\qquad
(\hat{a},\,\hat{\rho}_{t+H},\,\hat{v})
= \operatorname{Unpack}(\hat{F}),
\label{eq:slot_readout}
\end{equation}
where $\operatorname{Unpack}$ selects the action, future-proprioception, and value slots
and restores their original shapes; task and current-proprioception bands serve as inputs.
The selected coefficients directly represent the normalized signals, so no further inverse
DCT is applied.}

\subsection{Conditioning-Only Input Meta-Frame}
The input meta-frame $z^{m}{\mathrm{in}}$ reuses the tokenizer's mandatory blank slot rather than a dedicated proprioceptive frame. It follows the same channel-band layout as
$z^{m}{\mathrm{out}}$, with task type and current proprioception $\rho_t$ always provided as
mandatory conditioning inputs. Other bands are task-dependent: the slot-aware mask in Sec.~\ref{subsec:routing} decides whether they serve as clean conditioning signals or denoising targets.
Since the tokenizer already reserves a leading blank slot, reusing this otherwise-idle slot makes input-side conditioning effectively free.

\subsection{Slot-Aware Task Routing}
\label{subsec:routing}
MetaWAM trains policy, world-model, and value prediction in one denoising model. The three tasks
share the same robot variables and differ only in causal direction: policy predicts the action
from the current state; the world model predicts the future state from state and action; value
estimation predicts the return from the full transition. The variables never move between
frames; they always occupy the same channel bands of the output meta-frame, and only their role changes per task.
We implement this with slot-aware masks. Let $m_{\pi}, m_{w}, m_{v}$ indicate policy,
world-model, and value samples; for each sample a band-level mask is defined as $M_{\mathrm{slot}} = \Phi(m_{\pi}, m_{w}, m_{v})$,
where $\Phi$ deterministically opens a band (clean conditioning) or closes it (denoising
target). The mask switches over the partition of
Eq.~\eqref{eq:channel_bands}, so each signal is switched as an atomic unit. The routing is shown
in Fig.~\ref{fig:framework} (bottom left): a policy sample opens only the task and
current-proprioception bands; a world-model sample additionally opens the action band, with the
future state still a target; a value sample further opens the future-proprioception band,
leaving only value as a target. Such band-level routing is strictly finer than a frame-level
mask, which could not open the action band while hiding the co-located value band.

\textbf{Dual-granularity masking.}
The routing runs at two granularities. The slot-level mask carries all task
semantics on the meta-frame; from it we deterministically derive a single-channel coarse mask
for the original video-denoising path, so the six visual frames are conditioned through exactly
the interface the pretrained VDM expects. The fine mask carries all task information; the coarse
mask carries none and exists only to keep byte-level compatibility with the backbone.

\subsection{Training Objective}
\label{subsec:objective}
Training follows the standard diffusion denoising objective over the full latent sequence. Given
a clean sequence $\mathbf{z}_0$, we sample noise $\epsilon$ and level $\sigma$ and train the
denoiser $D_{\theta}$ under language and video conditioning:
\begin{equation}
\mathcal{L}_{\mathrm{diff}}
=
\mathbb{E}_{\mathbf{z}_0,\epsilon,\sigma}
\!\left[
w(\sigma)\,
\big\|
D_{\theta}(\mathbf{z}_0+\sigma\epsilon,\sigma,c,M) - \mathbf{z}_0
\big\|_2^2
\right].
\label{eq:diffusion_loss}
\end{equation}
{We additionally supervise the robot signals recovered from the predicted output meta-frame
using Eq.~\eqref{eq:slot_readout}:
\begin{equation}
\begin{aligned}
\mathcal{L}
={}& \mathcal{L}_{\mathrm{diff}}
+ \lambda_{\mathrm{act}}\mathcal{L}_{\mathrm{act}}
+ \lambda_{\mathrm{prop}}\mathcal{L}_{\mathrm{prop}} \\
&+ \lambda_{\mathrm{val}}\mathcal{L}_{\mathrm{val}}
+ \lambda_{\mathrm{off}}\mathcal{L}_{\mathrm{off}}.
\end{aligned}
\label{eq:total_loss}
\end{equation}
The three signal terms are Huber losses ($\delta=1$), averaged over signal elements and
eligible samples: policy samples for action, policy/world samples for future proprioception,
and all tasks for value. Targets are read from the ground-truth meta-frame and directly
represent normalized signals. Coefficients outside the allocated signal slots are penalized by
\begin{equation}
\mathcal{L}_{\mathrm{off}}
=
\mathbb{E}\!\left[
\operatorname{mean}_{C,H,W}
\left|(1-S)\odot\hat{F}\right|
\right],
\label{eq:offslot_loss}
\end{equation}
where $S$ marks allocated coefficient positions, the expectation covers all tasks, and
$\lambda_{\mathrm{off}}=10^{-4}$. DCT and slot readout remain differentiable, so all
terms backpropagate to the denoiser without a learned decoding head.}

\section{Experiments}
\label{sec:result}
We evaluate MetaWAM on RoboCasa~\citep{nasiriany2024robocasa}, LIBERO~\citep{liu2023libero},
LIBERO-Plus~\citep{fei2025liberoplus}, and three real-world Franka tabletop tasks.
RoboCasa covers 24 static kitchen tasks with 50 human demonstrations per task; LIBERO covers
four manipulation suites, and LIBERO-Plus tests robustness to seven perturbation categories.
We initialize Cosmos-Predict2-2B-Video2World~\citep{cosmoswfm2025}. For RoboCasa, we use
AdamW with learning rate $10^{-6}$, batch size 64 per GPU, and four gradient-accumulation
steps on four NVIDIA H200 GPUs.

\begin{wraptable}{r}{0.45\textwidth}
\centering
\vspace{-30pt}
\small
\setlength{\tabcolsep}{4pt}
\renewcommand{\arraystretch}{1.08}
\begin{tabular}{lcc}
\toprule
\textbf{Method} & \makecell[c]{\textbf{Demos} \\ \textbf{/ Task}} & \makecell[c]{\textbf{Avg.} \\ \textbf{SR (\%)}} \\
\midrule
UVA~\citep{uva2025} & 50 & 50.0 \\
DP-VLA~\citep{han2024dual} & 3000 & 57.3 \\
UWM~\citep{uwm2025} & 1000 & 60.8 \\
$\pi_0$~\citep{pi02024} & 300 & 62.5 \\
GR00T-N1.5~\citep{bjorck2025gr00t} & 300 & 64.1 \\
Video Policy~\citep{liang2025video} & 300 & 66.0 \\
FLARE~\citep{zheng2025flare} & 300 & 66.4 \\
Cosmos Policy~\citep{cosmospolicy2026} & 50 & 67.1 \\
\oursrow \textbf{MetaWAM (ours)} & \textbf{50} & \textbf{68.9} \\
\bottomrule
\end{tabular}
\vspace{-5pt}
\caption{RoboCasa benchmark results.}
\label{tab:robocasa}
\vspace{-12pt}
\end{wraptable}

\subsection{Evaluation in Simulation Environments}
\textbf{RoboCasa.}
We first evaluate MetaWAM on RoboCasa under the low-demonstration regime. Following the
Cosmos Policy evaluation protocol, each task is evaluated over 50 trials across five evaluation
scenes with different floor plans and visual styles, with 10 trials per scene; we report the
average success rate across all 24 tasks and 3 random seeds, totaling 3600 trials. As shown in
Tab.~\ref{tab:robocasa}, MetaWAM reaches 68.9\% average success using only 50 demonstrations
per task, outperforming the frame-injection Cosmos Policy baseline under the same data setting
and comparing favorably with baselines trained on far more demonstrations. The unified
meta-frame interface thus improves policy performance while collapsing the per-signal frames
into a single carrier, rather than trading accuracy for compactness.

\textbf{LIBERO and LIBERO-Plus.}
MetaWAM achieves 98.1\% average success across the Spatial, Object, Goal, and Long suites,
reported in the Original column of Table~\ref{tab:libero_plus}. Training and evaluation
protocols are provided in the supplementary material.
To examine performance beyond the standard benchmark, we evaluate seven perturbation
categories: camera viewpoint, robot initial state, language instruction, lighting, background,
sensor noise, and object layout. Table~\ref{tab:libero_plus} compares our results with results from~\citep{zhang2026wamrobustness}.
MetaWAM reaches 83.44\% overall success, exceeding the reported Cosmos Policy result
of 82.2\% by 1.24 percentage points and attaining the highest overall score among the listed
WAMs. Its camera and language scores reach 82.93\% and 89.56\%, improving on Cosmos
Policy by 7.13 and 7.86 points, respectively. It also leads the listed methods under lighting
(97.81\%), sensor noise (94.87\%), and layout (86.73\%) perturbations.
These gains span five of the seven categories, although robustness remains uneven:
background (72.68\%) and robot initial-state changes (59.90\%) are weaker than Cosmos
Policy. The comparison highlights
both the robustness of the compact interface across several perturbations and the remaining
sensitivity to visual context and initial robot configuration.

\begin{table}[!ht]
\centering
\footnotesize
\setlength{\tabcolsep}{2.7pt}
\renewcommand{\arraystretch}{1.0}
\begin{tabular}{lrrrrrrrrr}
\toprule
\textbf{Model} & \textbf{Original} & \textbf{Camera} & \textbf{Robot} & \textbf{Lang.} & \textbf{Light} & \textbf{BG} & \textbf{Noise} & \textbf{Layout} & \textbf{Total} \\
\midrule
\multicolumn{10}{c}{\textit{VLAs}} \\
$\pi_0$~\citep{pi02024} & 94.2 & 13.8 & 6.0 & 58.8 & 85.0 & 81.4 & 79.0 & 68.9 & 53.6 \\
$\pi_0$-FAST~\citep{fast2025} & 85.5 & 65.1 & 21.6 & 61.0 & 73.2 & 73.2 & 74.4 & 68.8 & 61.6 \\
OpenVLA-OFT~\citep{kim2025fine} & 97.6 & 55.6 & 21.7 & 81.0 & 92.7 & 91.0 & 78.6 & 68.7 & 67.9 \\
UniVLA~\citep{bu2025univla} & 95.2 & 1.8 & 46.2 & 69.6 & 69.0 & 81.0 & 21.2 & 31.9 & 42.9 \\
RIPT-VLA~\citep{tan2025ript} & 97.5 & 55.2 & 31.2 & 77.6 & 88.4 & 91.6 & 73.5 & 74.2 & 68.4 \\
X-VLA~\citep{zheng2025xvla} & 98.1 & 23.4 & \textbf{89.7} & 75.7 & 88.2 & \textbf{96.0} & 62.7 & 71.8 & 71.4 \\
HoloBrain0-GD~\citep{lin2026holobrain} & 96.7 & 65.5 & 58.2 & 78.7 & 88.1 & 90.3 & 66.9 & 79.5 & 74.0 \\
ABot-M0~\citep{yang2026abot} & \textbf{98.6} & 60.4 & 67.9 & 86.4 & 96.2 & 91.6 & 86.4 & 82.6 & 80.5 \\
\midrule
\multicolumn{10}{c}{\textit{VLA + WM}} \\
VLA-JEPA~\citep{sun2026vlajepa} & 97.2 & 64.2 & 67.7 & 88.1 & 91.8 & 93.4 & 65.8 & 83.9 & 77.9 \\
\midrule
\multicolumn{10}{c}{\textit{WAMs}} \\
GE-Act~\citep{liao2025genie} & 94.4 & 60.7 & 77.0 & 77.4 & 95.8 & 86.0 & 90.9 & 80.2 & 80.3 \\
Cosmos Policy~\citep{cosmospolicy2026} & 98.5 & 75.8 & 63.3 & 81.7 & 96.5 & 88.9 & 92.7 & 82.2 & 82.2 \\
Fast-WAM~\citep{fastwam2026} & 97.6 & 16.4 & 44.5 & 68.9 & 78.2 & 53.7 & 37.7 & 60.7 & 51.5 \\
\oursrow MetaWAM (ours) & 98.1 & \textbf{82.9} & 59.9 & \textbf{89.5} & \textbf{97.8} & 72.6 & \textbf{94.8} & \textbf{86.7} & \textbf{83.4} \\
\bottomrule
\end{tabular}
\caption{LIBERO-Plus success rates (\%). Baselines are reported in Table~4 of \citep{zhang2026wamrobustness}; Original denotes standard LIBERO.
Lang.: language; BG: background. Bold marks the best result per column.}
\label{tab:libero_plus}
\end{table}

\begin{figure*}[!t]
\centering
    \includegraphics[width=0.85\linewidth]{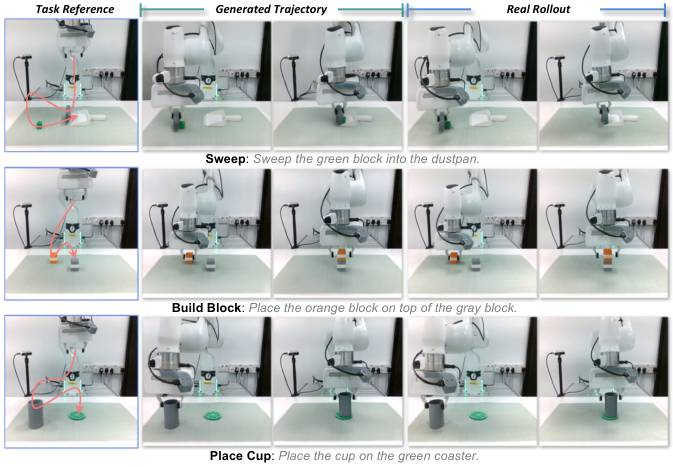}
    \caption{\textbf{Real-world rollouts.} For each of the three real-world tasks, we show a task
    reference, a representative generated trajectory, and a representative real rollout.}
    \label{fig:qualitative_result}
\centering
\end{figure*}

\subsection{Real-World Experiments}
We evaluate three precision-sensitive tasks with a 7-DoF Franka Panda: Sweep moves a block
into a dustpan, Place Cup aligns a cup with a coaster, and Build Block stacks cubes.
The OOD setting adds distractors, unseen object placements, and background changes.
MetaWAM reaches 85.0\% ID and 68.3\% OOD success, improving over Cosmos Policy in
both settings (Table~\ref{tab:real-world}); Fig.~\ref{fig:qualitative_result} shows representative rollouts.

\begin{table}[htbp]
    \centering
    \small
    \renewcommand{\arraystretch}{1.2}
    \setlength{\tabcolsep}{4pt}
    \resizebox{\linewidth}{!}{
    \begin{tabular}{l cccc c cccc}
        \toprule
        \multirow{2}{*}{\textbf{Method}} & \multicolumn{4}{c}{\textbf{In-Distribution (ID)}} & & \multicolumn{4}{c}{\textbf{Out-of-Distribution (OOD)}} \\
        \cmidrule{2-5} \cmidrule{7-10}
        & Sweep & Place Cup & Build Block & \textbf{SR\%} & & Sweep & Place Cup & Build Block & \textbf{SR\%} \\
        \midrule
        $\pi_0$~\citep{pi02024}  & 45.0 & 50.0 & 55.0 & 50.0 & & 35.0 & 35.0 & 40.0 & 36.7 \\
        $\pi_{0.5}$~\citep{intelligence2025pi_}  & 70.0 & 70.0 & 75.0 & 71.7 & & 45.0 & 45.0 & 50.0 & 46.7 \\
        Cosmos Policy~\citep{cosmospolicy2026} & 75.0 & 80.0 & 80.0 & 78.3 & & 60.0 & 60.0 & 70.0 & 63.3 \\
        \oursrow \textbf{MetaWAM (Ours)} & \textbf{80.0} & \textbf{85.0} & \textbf{90.0} & \textbf{85.0} & & \textbf{65.0} & \textbf{65.0} & \textbf{75.0} & \textbf{68.3} \\
        \bottomrule
    \end{tabular}
    }
    \caption{Real-world success rates (\%) over 20 trials per task under ID and OOD settings.}
    \label{tab:real-world}
\end{table}

\subsection{Ablation Study}
All ablations are trained and evaluated on RoboCasa under the same 50-demonstration-per-task
setting and evaluation protocol as the main simulation experiments. We report the average
success rate (SR) across the 24 RoboCasa tasks.

\textbf{Robot-signal encoding.}
Table~\ref{tab:encoding_ablation} compares success and inference latency.
Separate signal frames require 0.479\,s per query; all compact interfaces reduce this to
0.291\,s. Our frequency-domain interface achieves the highest success (68.9\%), exceeding
the spatial quadrant and compact-region packings at the same latency.

\begin{table}[t]
\centering
\small
\setlength{\tabcolsep}{2.8pt}
\renewcommand{\arraystretch}{1.1}
\begin{tabular}{lcccccccc}
\toprule
& & \textbf{SR} & \textbf{Latency}
& \multicolumn{2}{c}{\textbf{Signal recovery}} & \multicolumn{3}{c}{\textbf{World/value prediction}} \\
\cmidrule(lr){5-6} \cmidrule(lr){7-9}
\textbf{Encoding} & \textbf{Fr.} & \textbf{(\%)}$\uparrow$ & \textbf{(s)}$\downarrow$
& \textbf{Gain}$\uparrow$ & \textbf{Rel.\ $L_2$}$\downarrow$
& \textbf{PSNR}$\uparrow$ & \textbf{Prop.\ NEE}$\downarrow$ & \textbf{Value MAE}$\downarrow$ \\
\midrule
Separate Spatial Dup.   & 11 & 67.1 & 0.479 & 0.6361 & 0.3657 & 19.27 & 0.0250 & 0.072 \\
\midrule
Meta-Quadrants          & 8  & 65.9 & 0.291 & 0.6032 & 0.3988 & 18.86 & 0.0499 & 0.081 \\
Meta-Compact            & 8  & 68.1 & 0.291 & 0.6563 & 0.3345 & 19.41 & 0.0193 & 0.065 \\
\oursrow Unified Meta-Frame & 8 & \textbf{68.9} & \textbf{0.291} & \textbf{0.7146} & \textbf{0.2928}
& \textbf{19.95} & \textbf{0.0143} & \textbf{0.046} \\
\bottomrule
\end{tabular}
\caption{Ablation on robot-signal encoding. Fr.: latent frames per sequence. Signal recovery
is measured with the frozen Cosmos-Predict2 denoiser of Tab.~\ref{tab:vdm_prior_compat};
world/value prediction is evaluated with the fine-tuned models. Prop.\ NEE: normalized error
of the predicted future proprioception.}
\label{tab:encoding_ablation}
\end{table}

\textbf{Signal recovery after denoising.}
A small latent correction does not by itself show that the robot signal survives denoising.
To test this, we decode the robot signal from the frozen-denoiser outputs of
Sec.~\ref{sec:motivating_analysis}, using only the encoded carrier, and report two
signal-space metrics averaged over the five noise levels: \emph{target-direction gain}, the
amplitude retained along the clean-signal direction, and \emph{relative readout $L_2$}, the
normalized error of the decoded signal. Both rank the four layouts in the same order as SR,
with our unified meta-frame best (Tab.~\ref{tab:encoding_ablation}). This agreement supports
the diagnostic as a comparative measure of frozen-VDM compatibility; we do not use it as a
quantitative predictor of SR.

\textbf{World and value prediction.}
Policy gains do not come at the expense of the other tasks: among the four layouts, the
unified meta-frame also attains the highest future-view PSNR and the lowest
future-proprioception NEE and value MAE (Tab.~\ref{tab:encoding_ablation}).

\vspace{-4pt}
\section{Conclusion}
\label{sec:conclusion}
MetaWAM treats robot-signal injection as a distribution-alignment problem.
Frozen-denoiser diagnostics favor frequency-domain packing over spatial duplication and
region-based alternatives. Fixed coefficient slots and analytical DCT readout provide a compact
interface for joint policy, world-model, and value learning. Results on RoboCasa, LIBERO,
LIBERO-Plus, and real-world Franka tasks demonstrate the interface's efficiency and applicability.

\textbf{Limitations.}
The fixed band layout and loss weights may require adaptation when task importance or
signal scales vary substantially. Instruction following remains limited by the
Cosmos-Predict2 backbone~\citep{cosmoswfm2025}; stronger language grounding is a useful
direction. LIBERO-Plus also exposes sensitivity to background and robot initial-state changes.

\textbf{Acknowledgment.}
Guoqiang Liang and Yiqi Lin were supported by the NUS Guangzhou Research Translation and Innovation Institute (GRTII)-Affiliated PhD Scholarship Program.

\clearpage

\bibliography{example}

\clearpage

\appendix

\section{More Details of Motivation Analysis}
\label{sec:suppl_motivation}

\subsection{Experiment Details}
\label{sec:suppl_motivation_experiment}

All motivating-analysis experiments are conducted using samples from the RoboCasa dataset~\citep{nasiriany2024robocasa} released with Cosmos Policy~\citep{cosmospolicy2026}.
For the single-signal analysis, we randomly select 30 action chunks from distinct episodes and compare different ways of injecting this low-dimensional signal into the latent space.
For the multi-signal packing analysis, we use the corresponding action chunks, proprioceptive states, and value labels to construct packed meta-frame variants, allowing us to evaluate the distributional compatibility of different multi-signal encoding schemes under the same pretrained video diffusion model.

We compare three spatial packing baselines against our frequency-domain meta-frame.
\emph{Sep.\ Spatial Dup.} follows the Cosmos Policy implementation, where different robot-related signals are injected into separate latent frames by spatially duplicating each signal across the latent grid.
\emph{Meta-Quadrants} packs multiple signals into a single latent frame by assigning each signal to an equally sized spatial region, and then replicating the signal values along the channel dimension within its assigned region.
\emph{Meta-Compact} also uses a single latent frame but allocates spatial regions according to signal dimensionality: longer signals, such as action chunks, occupy larger regions, while shorter signals, such as proprioception and value, use smaller regions.
As in Meta-Quadrants, the assigned signal values are replicated along the channel dimension within each spatial region.

We evaluate five diffusion noise levels, and for each noise level $\sigma$, we repeat the denoising evaluation five times with independently sampled noise and report the average across trials.

\subsection{Cross-Backbone Analysis on Wan2.1 and Wan2.2}
\label{sec:suppl_wan_analysis}

To further examine whether the observed advantage of frequency-domain encoding is specific to Cosmos-Predict2~\citep{cosmoswfm2025}, we conduct the same motivating analysis on two additional video diffusion backbones from the Wan series~\citep{wan2025}: Wan2.1-I2V-14B and Wan2.2-TI2V-5B.
As shown in Tab.~\ref{tab:wan21} and Tab.~\ref{tab:wan22}, DCT Encoding consistently achieves lower denoiser reconstruction error than spatial duplication and VAE-based encoding in the single-signal injection setting, while our Unified Meta-Frame outperforms all spatial packing baselines in the multi-signal packing setting.
These results demonstrate that the benefit of frequency-domain signal injection is not tied to a particular model size, video-diffusion backbone, or training objective.
Specifically, the same trend holds across Wan2.1-I2V-14B and Wan2.2-TI2V-5B, which are trained with flow matching, as well as Cosmos-Predict2, which follows an EDM-style residual denoising formulation.
This cross-backbone consistency further supports our claim that DCT-based meta-frames provide a more VDM-compatible interface for injecting low-dimensional robot signals into pretrained video diffusion models.

\begin{table}[!ht]
\centering
\begin{minipage}[t]{0.49\linewidth}
\centering
\resizebox{\linewidth}{!}{%
\setlength{\tabcolsep}{3pt}
\renewcommand{\arraystretch}{1.1}
\begin{tabular}{lccccc}
\toprule
\textbf{Method} & 0.2 & 0.4 & 0.6 & 0.8 & 0.9 \\
\midrule
Spatial Duplication & 0.8858 & 0.9883 & 1.0092 & 1.0434 & 1.0951 \\
VAE Encoding        & \textbf{0.7534} & 0.8993 & 0.9891 & 1.0478 & 1.0989 \\
\oursrow DCT Encoding        & 0.7701 & \textbf{0.8258} & \textbf{0.9053} & \textbf{0.9943} & \textbf{1.0902} \\
\bottomrule
\end{tabular}%
}
\caption*{{(a) Single-signal injection.}}
\end{minipage}
\hfill
\begin{minipage}[t]{0.49\linewidth}
\centering
\resizebox{\linewidth}{!}{%
\setlength{\tabcolsep}{3pt}
\renewcommand{\arraystretch}{1.1}
\begin{tabular}{lccccc}
\toprule
\textbf{Method} & 0.2 & 0.4 & 0.6 & 0.8 & 0.9 \\
\midrule
Sep.\ Spatial Dup.$^\dagger$ & 0.9029 & 0.9944 & 1.0194 & 1.0519 & 1.1004 \\
Meta-Quadrants               & 0.9182 & 1.0005 & 1.0271 & 1.0580 & 1.1047 \\
Meta-Compact                 & 0.8467 & 0.9990 & 1.0226 & 1.0540 & 1.1024 \\
\oursrow Unified Meta-Frame  & \textbf{0.7144} & \textbf{0.8710} & \textbf{0.9505} & \textbf{1.0226} & \textbf{1.0988} \\
\bottomrule
\end{tabular}%
}
\caption*{{(b) Multi-signal packing.}}
\end{minipage}
\vspace{-5pt}
\caption{Wan2.1-I2V-14B VDM-prior compatibility, measured by the relative reconstruction error of the frozen denoiser at noise level $\sigma$ ($\downarrow$ lower is better).
\textbf{Bold} indicates the best method at each $\sigma$ among the listed methods.
$^\dagger$Sep.\ Spatial Dup. averages the reconstruction error over separate action, proprioception, and value latent frames.}
\vspace{-10pt}
\label{tab:wan21}
\end{table}

\begin{table}[!ht]
\centering
\begin{minipage}[t]{0.49\linewidth}
\centering
\resizebox{\linewidth}{!}{%
\setlength{\tabcolsep}{4pt}
\renewcommand{\arraystretch}{1.1}
\begin{tabular}{lccccc}
\toprule
\textbf{Method} & 0.2 & 0.4 & 0.6 & 0.8 & 0.9 \\
\midrule
Spatial Duplication    & 0.9777 & 1.0706 & 1.1058 & 1.1566 & 1.2187 \\
VAE Encoding           & 0.7537 & 1.0241 & 1.0659 & 1.1025 & 1.1445 \\
\oursrow DCT Encoding  & \textbf{0.7059} & \textbf{0.8662} & \textbf{0.9572} & \textbf{1.0419} & \textbf{1.0992} \\
\bottomrule
\end{tabular}%
}
\caption*{{(a) Single-signal injection.}}
\end{minipage}
\hfill
\begin{minipage}[t]{0.49\linewidth}
\centering
\resizebox{\linewidth}{!}{%
\setlength{\tabcolsep}{4pt}
\renewcommand{\arraystretch}{1.1}
\begin{tabular}{lccccc}
\toprule
\textbf{Method} & 0.2 & 0.4 & 0.6 & 0.8 & 0.9 \\
\midrule
Sep.\ Spatial Dup.$^\dagger$ & 0.9942 & 1.0768 & 1.1204 & 1.1748 & 1.2603 \\
Meta-Quadrants               & 1.0030 & 1.0814 & 1.1188 & 1.1683 & 1.2482 \\
Meta-Compact                 & 1.0461 & 1.1625 & 1.2384 & 1.3167 & 1.4663 \\
\oursrow Unified Meta-Frame  & \textbf{0.7647} & \textbf{0.9213} & \textbf{1.0011} & \textbf{1.0683} & \textbf{1.1307} \\
\bottomrule
\end{tabular}%
}
\caption*{{(b) Multi-signal packing.}}
\end{minipage}
\vspace{-5pt}
\caption{Wan2.2-TI2V-5B VDM-prior compatibility, measured by the relative reconstruction error of the frozen denoiser at noise level $\sigma$ ($\downarrow$ lower is better).
\textbf{Bold} indicates the best method at each $\sigma$ among the listed methods.
$^\dagger$Sep.\ Spatial Dup. averages the reconstruction error over separate action, proprioception, and value latent frames.}
\vspace{-20pt}
\label{tab:wan22}
\end{table}

\section{Additional Experimental Details}
\label{sec:suppl_experiment}

\subsection{LIBERO Simulation Benchmark}
\label{sec:suppl_libero}

\begin{table}[!ht]
\centering
\small
\setlength{\tabcolsep}{5pt}
\begin{tabular}{lccccc}
\toprule
\textbf{Method} & \textbf{Spatial} & \textbf{Object} & \textbf{Goal} & \textbf{Long} & \textbf{Average} \\
\midrule
Diffusion Policy~\citep{diffusionpolicy2023} & 78.3 & 92.5 & 68.3 & 50.5 & 72.4 \\
Dita~\citep{hou2025dita}                     & 97.4 & 94.8 & 93.2 & 83.6 & 92.3 \\
$\pi_0$~\citep{pi02024}                      & 96.8 & 98.8 & 95.8 & 85.2 & 94.2 \\
UniVLA~\citep{bu2025univla}                  & 96.5 & 96.8 & 95.6 & 92.0 & 95.2 \\
$\pi_{0.5}$~\citep{intelligence2025pi_}       & 98.8 & 98.2 & 98.0 & 92.4 & 96.9 \\
OpenVLA-OFT~\citep{kim2025fine}              & 97.6 & 98.4 & 97.9 & 94.5 & 97.1 \\
CogVLA~\citep{li2026cogvla}                  & 98.6 & 98.8 & 96.6 & 95.4 & 97.4 \\
Cosmos Policy~\citep{cosmospolicy2026}        & 98.1 & \textbf{100.0} & \textbf{98.2} & \textbf{97.6} & \textbf{98.5} \\
MetaWAM (ours)                               & \textbf{99.0} & 99.8 & 97.4 & 96.0 & 98.1 \\
\bottomrule
\end{tabular}
\caption{LIBERO simulation benchmark results.}
\label{tab:libero_results}
\end{table}

The LIBERO benchmark~\citep{liu2023libero} includes four primary task suites: LIBERO-Spatial, LIBERO-Object, LIBERO-Goal, and LIBERO-Long. Each task suite provides a training dataset of 500 demonstrations in total, covering 10 tasks with 50 demonstrations per task. We follow the standard evaluation protocol used by prior works~\citep{cosmospolicy2026,kim2024openvla}. For each suite, success rates are averaged over 500 evaluation trials, corresponding to 10 tasks with 50 episodes per task. We evaluate each method with three random seeds, resulting in 6000 evaluation trials in total across the four LIBERO suites. We train MetaWAM on LIBERO for 12K optimization steps using 4 NVIDIA H200 GPUs with a global batch size of 1200, requiring 56 hours of wall-clock training time. The policy predicts an action chunk of 16 timesteps, and we execute all steps before re-querying the policy, following the setting of Cosmos Policy~\citep{cosmospolicy2026}.

\subsection{Training Details and Real-Robot Setup}
\label{sec:suppl_training}

\textbf{RoboCasa Training Details.}
We train MetaWAM on RoboCasa for 12K optimization steps using 4 NVIDIA H200 GPUs with a global batch size of 1024, requiring 64 hours of wall-clock training time.
All parameters of the base Cosmos-Predict2~\citep{cosmoswfm2025} model are fully fine-tuned.
The policy predicts an action chunk of 32 timesteps, and we execute 16 steps before re-querying the policy.

\begin{figure}[!ht]
    \centering
    \includegraphics[width=0.75\linewidth]{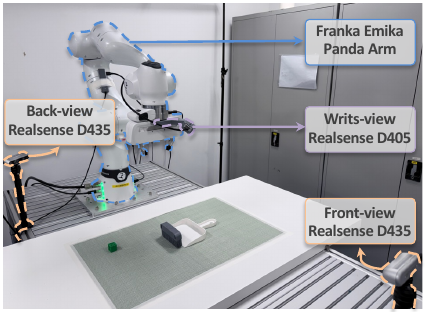}
    \caption{\textbf{Real-robot hardware setup.}
    The platform consists of a 7-DoF Franka Emika Panda arm, a wrist-mounted Intel RealSense D405 camera, and two external Intel RealSense D435 cameras placed at the front and back of the workspace.
    The wrist camera provides observations near the end-effector, whereas the two external cameras provide third-person views.}
    \vspace{-3pt}
    \label{fig:suppl_setup}
\end{figure}

\textbf{Real-World Franka Training Details.}
For the real-world Franka experiments, we train MetaWAM for 8K optimization steps using 4 NVIDIA H200 GPUs with a global batch size of 1024, requiring approximately 47 hours of wall-clock training time.
We use the same computational resources for all baselines, including $\pi_0$~\citep{pi02024}, $\pi_{0.5}$~\citep{intelligence2025pi_}, and Cosmos Policy~\citep{cosmospolicy2026}.
We train the policy with an action chunk size of 16 timesteps and execute the entire predicted chunk before re-querying.
For each of the three real-world tasks, Build Block, Place Cup, and Sweep, we collect 80 demonstration episodes.

\begin{figure*}[!t]
\centering
    \includegraphics[width=1\linewidth]{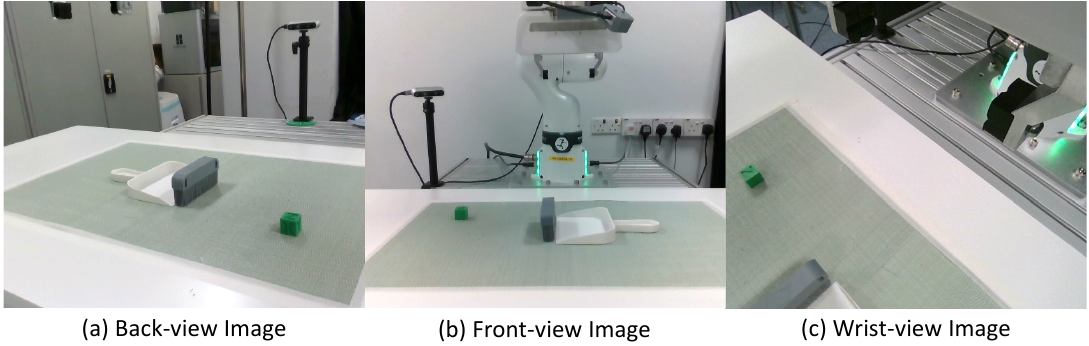}
    \vspace{-1mm}
    \caption{\textbf{Multi-view observations in the real-robot setup.}
    We use three RGB-D camera views for real-world policy execution:
    a back-view camera, a front-view camera, and a wrist-mounted camera.}
    \label{fig:suppl_view}
\vspace{-1mm}
\end{figure*}

\textbf{Real-Robot Setup.}
Our real-world experiments are conducted on a 7-DoF Franka Emika Panda manipulation platform with a three-view RGB-D perception system.
We use synchronized observations from the front-view, back-view, and wrist-view cameras as the visual input to the policy.
This design provides both global spatial context and local manipulation detail: the external views help localize objects and monitor the full manipulation scene, while the wrist view improves visibility around the gripper, especially when external cameras suffer from occlusions.
Fig.~\ref{fig:suppl_setup} illustrates the physical arrangement of the robot and cameras, and Fig.~\ref{fig:suppl_view} shows representative observations from the three camera views.
During inference, the model runs on an NVIDIA RTX 5090 GPU.

\subsection{More Visualization Results}
\label{sec:suppl_visualization}

\begin{figure*}[!t]
\centering
    \includegraphics[width=1\linewidth]{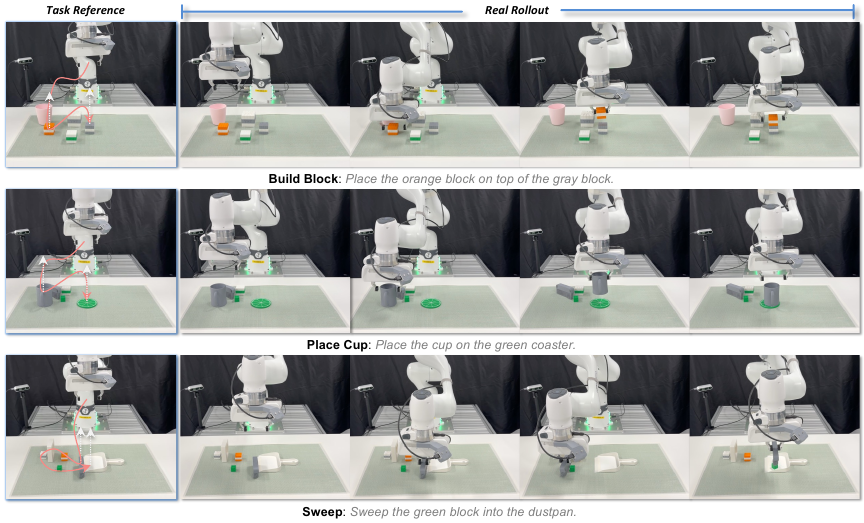}
    \vspace{-1mm}
    \caption{\textbf{Visualization results in real-world OOD scenarios.}
    We show representative key frames from real-robot rollouts under distribution shifts, including changed object locations, additional distractor objects, and background or layout variations.
    These examples complement the quantitative OOD evaluation and illustrate the robustness of MetaWAM under scene-level changes.}
    \label{fig:suppl_real_ood}
\vspace{-1mm}
\end{figure*}

Fig.~\ref{fig:suppl_real_ood} provides additional qualitative results for real-world out-of-distribution evaluation.
Compared with the in-distribution demonstrations, these rollouts introduce moderate scene shifts, including changed object locations, additional distractor objects, and background or layout variations.
These examples complement the quantitative real-world evaluation in the main paper by showing that MetaWAM can still produce coherent action trajectories under visual and spatial changes in the workspace.

\begin{figure*}[!t]
\centering
    \includegraphics[width=\linewidth]{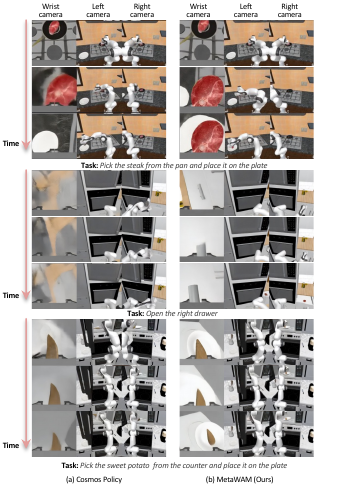}
    \vspace{-1mm}
    \caption{\textbf{Visualization results on RoboCasa.}
    We compare generated trajectories from Cosmos Policy and MetaWAM at a resolution of $224 \times 224$.
    The examples show that MetaWAM preserves coherent future-observation generation while using a compact unified meta-frame to encode robot-related signals.}
    \label{fig:suppl_robocasa}
\vspace{-1mm}
\end{figure*}

Fig.~\ref{fig:suppl_robocasa} further visualizes generated trajectories on the RoboCasa benchmark.
We compare trajectories generated by Cosmos Policy and MetaWAM at the same image resolution of $224 \times 224$.
These qualitative examples provide an intuitive view of the simulated rollouts used in our evaluation and complement the quantitative RoboCasa results reported in the main paper.
In particular, they show that the proposed unified meta-frame interface preserves the video-diffusion backbone's ability to generate temporally coherent future observations, while replacing separate robot-signal latent frames with a compact frequency-domain meta-frame.

\section{Discussion}
\label{sec:suppl_discussion}

\subsection{Attention to Encoded Robot Signals}
\label{sec:suppl_signal_attention}
\begin{figure}[!ht]
\centering
    \includegraphics[pagebox=cropbox,width=0.5\linewidth]{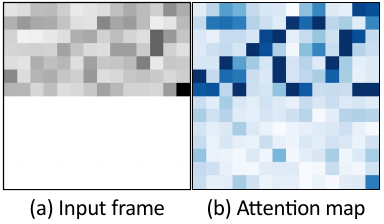}
    \caption{\textbf{Frozen-VDM attention to encoded robot signals.}
    The input carrier (left) and block-14 attention map (right) show an attention lift of
    $1.40\!\times$ over the encoded region relative to the full-frame mean. White regions
    in the input are padding.}
    \label{fig:suppl_signal_attention}
\end{figure}

The reconstruction diagnostic in Sec.~\ref{sec:motivating_analysis} measures how strongly
an injected representation is corrected by a frozen denoiser. We complement it by examining
whether the frozen VDM allocates attention to the encoded robot-signal region.
Figure~\ref{fig:suppl_signal_attention} visualizes attention at block 14 alongside the input
carrier; the white input region is padding. We define \emph{attention lift} as the mean
attention over the encoded region divided by the mean attention over the entire frame.
The visualized case has a lift of $1.40\!\times$, indicating above-average attention to the
encoded region. This observation is inconsistent with the region simply being ignored as
smooth background. Attention alone does not establish successful signal recovery or causal
use, but provides complementary evidence to the frozen-denoiser reconstruction analysis.

\subsection{Analytical Readout and Band Supervision}
\label{sec:suppl_analytical_interface}
\begin{figure}[!ht]
\centering
    \includegraphics[pagebox=cropbox,width=0.95\linewidth]{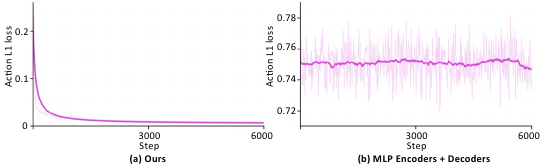}
    \caption{\textbf{Action recovery with analytical and learned interfaces.}
    Training action-chunk MAE for MetaWAM (left) and the MLP encoder/decoder control
    (right). Action $L_1$ is a monitoring metric, not an additional loss. The two panels
    use different vertical scales.}
    \label{fig:suppl_interface_convergence}
\end{figure}

\textbf{Comparison with a learned interface.}
We compare the analytical interface with a learned-projection control based on the
state/action interface of DreamZero~\citep{dreamzero2026}, extended to value using MLP
encoders and decoders for all three signals. Figure~\ref{fig:suppl_interface_convergence}
reports action-chunk mean absolute error (action $L_1$) during training. This quantity is
used solely for monitoring and is not an additional training objective. Under the tested
setup, MetaWAM reduces this error rapidly, while the learned MLP interface remains at a
high error level.

In the unified policy, world-model, and value setting, a robot signal can serve as a clean
condition or a prediction target depending on the task. A learned interface therefore needs
per-signal encoding and decoding together with mask-aware routing; updating its encoders
also changes the latent targets presented to the denoiser. A zero-initialized projection
initially conveys no signal. Fixed packing and DCT/iDCT instead provide a static analytical
mapping and readout from the start of training. The observed optimization behavior supports
this design for the tested setup, without implying that learned interfaces cannot work with
other architectures or training procedures.

\end{document}